\documentclass{article}
\usepackage{ijcai24}

\usepackage{times}
\usepackage{soul}
\usepackage{url}
\usepackage[hidelinks]{hyperref}
\usepackage[utf8]{inputenc}
\usepackage[small]{caption}
\usepackage{graphicx}
\usepackage{amsmath}
\usepackage{amssymb}
\usepackage{amsthm}
\usepackage{booktabs}
\usepackage{algorithm}
\usepackage[switch]{lineno}
\usepackage[noend]{algpseudocode}
\usepackage{multirow}
\usepackage{array}
\usepackage{xcolor}
\usepackage{colortbl}
\usepackage{bm}

\usepackage[inline]{enumitem}
\usepackage{natbib}
\newtheorem{challenge}{Vision}
\newtheorem{example}{Running Example}
\newtheorem{definition}{Definition}
\title{Towards Foundation-model-based Multiagent System\\to Accelerate AI for Social Impact}

\author{Yunfan Zhao
$^{1,2}$ \and Niclas Boehmer$^{1,3}$ \and Aparna Taneja$^4$ \And Milind Tambe$^{1,4}$\\
\affiliations
$^1$Harvard University, $^2$ GE Healthcare, $^3$Hasso Plattner Institute,
$^4$Google Deepmind}

\begin{document}

\maketitle

\begin{abstract}
AI for social impact (AI4SI) offers significant potential for addressing complex societal challenges in areas such as public health, agriculture, education, conservation, and public safety. However, existing AI4SI research is often labor-intensive and resource-demanding, limiting its accessibility and scalability; the standard approach is to design a (base-level) system tailored to a specific AI4SI problem.
We propose the development of a novel meta-level multi-agent system designed to accelerate the development of such base-level systems, thereby reducing the computational cost and the burden on social impact domain experts and AI researchers. Leveraging advancements in foundation models and large language models, our proposed approach focuses on resource allocation problems providing help across the full AI4SI pipeline from problem formulation over solution design to impact evaluation. We highlight the ethical considerations and challenges inherent in deploying such systems and emphasize the importance of a human-in-the-loop approach to ensure the responsible and effective application of AI systems.
\end{abstract}

% ==== INTRO ====
\section{Introduction}
Artificial intelligence (AI) for social impact (AI4SI), which focuses on leveraging AI to address societal issues, has gained traction in both academia and industry \cite{bondi2021envisioning, DBLP:journals/aim/PerraultFST20,cowls2021definition,10628639,foffano2023investing}. With advancements in AI and multi-agent systems, there is an opportunity to apply these technologies to complex problems in areas like public safety, wildlife conservation, and public health  \cite{kwok2019ai, wang2019ai, floridi2018ai4people, ji2024mitigating}.
Previously, AI4SI research has been very labor-intensive, as it is oftentimes necessary to develop customized approaches going beyond conventional methods to address the challenges characteristic to these domains such as low resources and noisy or scarce data. This limits the overall impact of AI4SI research, as every individual effort requires non-trivial time, expertise, and financial investment. We envision the formation of a new approach to AI4SI which is less labor-intensive, customizable by non-experts, and can thus be made more widely available.
We believe promising progress can be made on this vision by employing recent methodological advancements in computer science research, tapping into a substantial currently unleveraged potential. %In the following, we will discuss directions of advances in computer science methods that we deem particularly relevant to AI4SI research, including multi-agent systems, foundation models, and Large Language Models. We will argue how these methods can accelerate AI4SI throughout the full AI4SI research pipeline.

In this paper, we will use resource allocation problems, which often arise in AI4SI domains \cite{kruk2018high, ayer2019prioritizing, DBLP:journals/aim/PerraultFST20, wang2019ai,baltussen2006priority,nishtala2021selective, deo2013improving} as our running example, yet our general ideas also apply more broadly. Some examples of previously studied resource allocation problems in AI4SI include strategically scheduling patrols in protected conservation areas \cite{gatmiry2021food, golovin2011dynamic, gordon2024combining,xu2016playing} and distributing scarce healthcare resources to optimize people's health outcomes \cite{shaikh2020artificial, mizan2022medical, vermagroup,zhao2024towards}. 
% where assigning patrol tasks to conservation agency personnel, we need to ensure that we strategically allocate limited staff so that important areas are covered and sufficiently patrolled \cite{gordon2024combining,xu2016playing}. 
% When distributing scarce healthcare resources to optimize people's health outcomes, we need to ensure that resources are given to people who would benefit substantially from them, while ensuring that underprivileged communities receive sufficient resources \cite{vermagroup,zhao2024towards}. 

{\bf We envision a meta-level multi-agent system that helps us accelerate the development of base-level systems, which tackle specific AI4SI problems.} The meta-level system would help non-profits and AI researchers in social impact domains leverage AI without having to invest significant amounts of labor and resources to build a tailored base-level system from scratch. 
% Specifically, we envision a multi-agent meta-level system where some agents help formulate the problem, while others design tailored solution methods for this problem and evaluate designed methods. Our envisioned meta-level system should not only help design a good initial base-level system but also iteratively adjust it based on human feedback. 
%We propose a meta-level multi-agent system to tackle AI for social impact problems. 
Our envisioned system leverages foundation models, which are typically developed by pretraining on available datasets, and can be used on different downstream tasks or new challenges \cite{bommasani2021opportunities, kenton2019bert, zhao2024aamas,viswanathan2023prompt2model}. Our proposed system involves: (i) using LLM based meta-level agents to communicate with decision makers in human language to understand the problem from the perspective of decision-makers; (ii) employing meta-level agents and foundation models for AI4SI problems to design base-level systems for AI4SI problems; and (iii) using meta-level agents for field-testing solutions to validate their impact. 

{\bf Importantly, instead of taking over the AI4SI pipeline, the meta-level system will accelerate the process, improve generalization, and enable thorough evaluation, with human-in-the-loop required in each part of the system.} We will discuss and highlight major challenges and our vision for each phase of this system, emphasizing the multi-agent aspect. We also discuss ethics and fairness aspects. 

The current era of foundation models has led to contemplation on the challenges and opportunities that such models may offer in multiple areas, from medical AI \cite{moor2023foundation} to autonomous supply chains \cite{xu2023multi}, autonomous mining \cite{10365454}, and robotics \cite{wang2024large}. In comparison with these previous works, this paper focuses on the use of foundation models in AI4SI. Moreover, even within AI4SI, we exemplarily focus on optimizing the allocation of limited resources, providing an analysis of challenges in a specific aspect of AI4SI. Furthermore, in contrast to previous work, we focus more on foundation-model-based agents at the meta-level, to assist base-level systems that optimize these limited resources, rather than replacing the base-level system entirely. This enables existing well-developed resource optimization tools to be brought to bear on relevant challenges as required, allowing instead the foundation models to configure the tools as needed.
\begin{figure*}[h!]
\centering
 \includegraphics[width=0.9\textwidth]{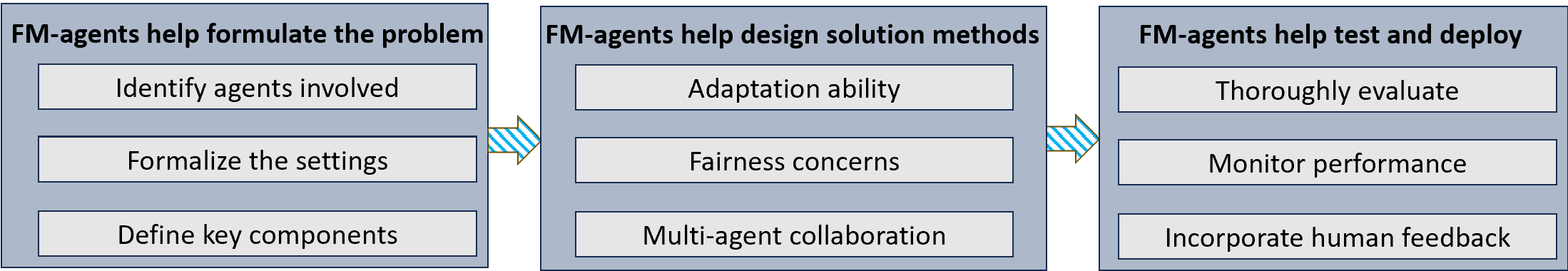}
\caption{Overview of our proposed AI for social impact (AI4SI) workflow. The three key phases, formulating the problem, designing the solution methods, and testing and deployment, are discussed in Sections~\ref{sec:phase1}, \ref{sec:phase2}, and \ref{sec:phase3} respectively. } 
\label{fig:overview_workflow}
\end{figure*}

\section{Preliminaries}
We formally define the key concepts of meta-level and base-level systems, which we will refer to throughout this work. 
\begin{definition}[Meta-level and base-level systems]
A base-level AI4SI system is the actual deployed system that will solve the problem on the ground. A meta-level system helps us accelerate the development of a base-level system or accelerate its modification as needed for a new objective. 
\end{definition}

Notably, a meta-level system does not actually solve an AI4SI problem. A meta-level system may involve several meta-level agents, each responsible for different tasks in the development of the base-level system, which may interact with each other. In the context of the use cases we are focusing on in this paper, the base system itself will be a multi-agent system or one that models multi-agent interactions such as a social network. The base-level agents present in the base-level system are defined as follows:
\begin{definition}[Base-level agents]
In the base-level system, the base-level agents model or serve as abstract representations of individuals or entities in the real-world. 
\end{definition}

Whereas the idea of a meta-level architecture has been proposed in agent architectures, there the idea is to directly assist agents in their immediate problem solving \cite{corkill1983use,rosenbloom1988meta,genesereth1983overview,wang2020m3rec}. In our work, the meta-level refers to deciding which agents and how many to select, how to evaluate their performance, and other such tasks. Another difference is that at least as conceived now, our meta-agents are focused on assisting humans in building base-level agents.

Another key concept we will frequently refer to is foundation-model-based agents (FM agents) used in the meta-level system.
\begin{definition}[Foundation-model-based agents (FM-agents)]
In the meta-level system, we use meta-level agents that employ foundation models including LLMs. We will refer to these agents as FM-agents
\end{definition}

An AI4SI pipeline typically involves three phases \cite{DBLP:journals/aim/PerraultFST20,bondi2021envisioning}: 
\begin{description}
    \item[Formulate the problem] Identify how to best represent or model the real-world entities and multiple stakeholders involved in the on-the-ground problem, along with their constraints and objectives. For example, the real-world entities may include individuals enrolled in social welfare programs and decision-makers such as non-profit program managers. Previously, this effort was largely manual and involved multiple parties, including nonprofits and AI researchers
    \cite{shi2020artificial,tomavsev2020ai,dou2022clinical}.
    \item[Design solution method] Identify appropriate solution methods and adapt them to design the base-level system. Previously, this process was largely manual in terms of researcher efforts to tailor advances in AI algorithms to the specific application scenarios. The adaptation to different application scenarios was also largely manual~\cite{wang2023scalable,dou2022learning,xiong2023reinforcement,wang2020guardhealth,paudel2022serologic}. 
    \item[Evaluation, refinement, and deployment] Testing, improving, and deploying solutions has typically been done manually, adding to the workload of researchers and even more importantly to that of resource-constraint  NGO workers \cite{wang2019ai}. 
\end{description}
Notably, each step in the AI4SI pipeline poses its own challenges \cite{sinha2018stackelberg}, implying that AI4SI work often requires substantially more effort than pure AI algorithmic improvement research. Specifically, formulating the problem and collecting detailed information (e.g. potential decisions available and their impact on each of the individuals) can be time-consuming and expensive \cite{sambasivan2022deskilling}. Moreover, designing solution methods can require significant time from AI experts and social impact domain experts \cite{tomavsev2020ai}, who must work together to devise a tailored solution method for every application scenario. Thorough testing and evaluation before deployment often require substantial manual effort, as detailed simulation studies are often required (e.g. by regulations or as precautionary measures) \cite{behari2024decision}, and AI experts typically manually design each simulation study from scratch. 

% pointed out that to address social challenges effectively with AI, it's crucial to first identify the agents involved in base-level systems and their decision-making processes. For example, understanding whether multiple agents, as in social networks, or just two, as in adversarial settings, are interacting is essential. In addition, to formally define a problem setting in commonly used terminologies in the AI community, we need to understand the range of actions available to the base-level agents and their impact on other base-level agents and the environment, such as the costs of these actions and the available budgets. 

% Throughout this work, to illustrate how one could potentially execute our proposed research ideas and visions, we will use as a running example the problem of 
For our running example, we consider allocating limited interventions (specifically live service calls made by health workers) in ARMMAN, which is a non-profit in India focusing on improving health awareness for expectant and new mothers \cite{zhao2024aamas,seow2024improving}. Their health
workers make service calls to boost the engagement of mothers enrolled in their health information program. They have shown that AI powered solutions can reduce engagement drops by about 30\% in real-world deployment \cite{mate2022field,wang2023scalable}. 
%Previous works have either modeled each beneficiary as a Markov Decision Process or modeled each beneficiary's trajectory as a time series \cite{danassis2023limited,seow2024improving}.  

\section{Formulating the Problem}
\label{sec:phase1}
In this phase, we discuss how to formulate an AI4SI problem.
%We start with describing existing approaches to formulate the problem, which are labor intensive. This motivates the use of FM-agents to help formulate the problem, potentially using the world knowledge of large language models. 
Existing works require researchers and human experts in AI to talk to collaborating non-profit organizations to understand who makes the decisions within a problem and gather key information on the agents such as demographics \cite{wang2019ai,wang2024healthq}. This process is expensive and labor-intensive, and non-profits may not have AI-trained staff to assist with this, making it difficult to ensure that AI solutions are accurately tailored to the complexities of the real-world problem. Thus, previous works often require AI researchers to have numerous rounds of discussions with non-profits and arrange a field trip to speak with key stakeholders in social good programs. Whereas these discussions are fundamental to AI4SI, some of the work oriented toward formulating the right base-level model is repetitive. To accelerate this work we propose the following vision:

%But this is often inconvenient due to language barriers, visa restrictions, and time conflicts. To overcome this, we propose the following vision: 

\begin{challenge}
Employ FM-agents to
\begin{enumerate*}[label=(\roman*)]
  \item 
identify the base-level agents involved,
  \item find an adequate formal description for the setting (e.g. as a Markov Decision Process), and 
  \item define key components of the settings (e.g. state space, action space, and reward function in a MDP).
\end{enumerate*} 
\end{challenge}

The FM-agents may use large language models to communicate with a partnering non-profit to formulate the social challenge as an AI problem and to understand who makes the decisions within a problem. The world knowledge of LLMs may help the FM-agent uncover confounding variables and undocumented information\cite{hu2024minicpm,wang2024efficient,he2023teacherlm,xu2024restful,ji2024reasoningrank,li2024focused}. The FM-agent should determine what information the individuals or entities in the real-world have to guide their decisions. The FM-agent should then define base-level agents to model the individuals or entities and make proper assumptions about the information available to them.

%such as assuming that an adversary base-level agent has insight into a defender base-level agent’s strategy. 

The FM-agent may gather data from past interactions, including the effects of actions, and observed costs or rewards, and interactions between agents \cite{roh2019survey,zhaoimplicit,elmachtoub2023estimate}. The FM-agent, potentially LLM-based, could communicate directly with beneficiaries enrolled in non-profit's program in their native language to gain a clearer perspective and would not have time constraints. %This would help figure out confounders and undocumented information which are critical to defining the formal problem setup. 

\begin{example}
In ARMMAN, the base-level agents represent beneficiaries enrolled, and the state and action space correspond to beneficiaries' engagement levels and possible schedules of service calls, respectively. The FM-agent should have conversations with domain experts in ARMMAN to find out that one way to approach this application scenario is to model beneficiaries as agents that follow Markov Decision Processes. After that, the FM-agent should define the state space, action space, and other key parts of the MDP. 
\end{example}

\section{Designing Solution Methods}
\label{sec:phase2}
In this phase, we elaborate on how to design a solution method for AI4SI  problems. Existing approaches often require manually designing solution methods tailored to each application scenario \cite{xu2022ranked,aqajari2024enhancing,mao2022towards, mao2022coem, Wang_EMBC2024_conerence,xiong2024whittle}. This approach fails to easily adapt to new application scenarios or knowledge and data from previous application scenarios, motivating the development of foundation models to accelerate solution approaches for problems in AI4SI. \textbf{Besides adaptation ability, other aspects that are of particular importance when designing solution methods in AI4SI problems include ethics, fairness, and collaboration among base-level agents. We will motivate each of these aspects and propose our visions to address them.}

% In designing this solution method, we leverage CS-methods advances in multi-agent systems, foundation models, and large language models to accommodate evolving needs of AI4SI domains. % and explicitly incorporate fairness, while mitigating the impact of noisy data. 

Adaptation is crucial in AI for social impact domains due to the dynamic and evolving nature of these environments. Social issues are often complex and multifaceted, with priorities evolving overtime \cite{blumenstock2018don, de2018machine, behari2024decision}. AI systems should adapt to new information and circumstances, accommodating rapidly changing needs and circumstances.
AI systems in social impact domains may serve diverse populations with varying needs, backgrounds, and experiences, highlighting the need for AI systems with strong adaptation and generalization abilities that can tailor their responses to specific subpopulations, ensuring inclusive decision making.

% A foundation model for resource allocation problems in AI4SI would allow us to build more generalizable systems, that could have broad applicability to different application scenarios, lower computational cost for each new application scenario, and better reach to underpreviledged communities \cite{bommasani2021opportunities,zhao2024aamas}. 
A foundation model for resource allocation could accelerate developing solutions for different application scenarios without incurring repeated development costs, making them more affordable and accessible to low-resource communities \cite{bommasani2021opportunities,zhao2024aamas}. For example, a foundation model designed to analyze medical data can be adapted to different diseases, health conditions, or populations, improving health outcomes on a larger scale \cite{li2023muben}. Based on advances in foundation models and adaptation, we propose the following vision:

\begin{challenge}
Build a foundation model for resource allocation problems in AI4SI domains that can be adapted to and finetuned on specific application scenarios. For each new AI4SI application scenario, employ a FM-agent to leverage the foundation model and provide solution methods. 
\end{challenge}

%To address the inherent complexity of social challenges, this FM-agent should employ Large Language Model (LLM) to process and understand human language and interpret instructions, queries, and feedback from stakeholders\cite{sun2024adaplanner,wang2024ecg,xiao2024configurable,xu2024can}. The FM-agent should incorporate human instructions to iteratively improve upon the solution method design for the base-level problem. The FM-agent should tailor the solution method to the need of the local community being served. 

\begin{example}
A concrete example of foundation model for resource allocations tasks is given by \citet{zhao2024towards}, who develop a pretrained restless bandit model that can be finetuned on various resource allocation application scenarios that ARMMAN may encounter. Currently, the application scenarios have different number of base-level agents and different amounts of distribution shifts. Here, our research idea is to start with such a foundation model and allow an FM-agent to adapt and specialize it to newer scenarios that may involve bigger changes than just differences in numbers of base-level agents. This could include new application scenarios that may need a change of the states and actions in the restless bandit model.
%Of course, An even more general model can potentially handle resource allocation tasks beyond restless bandits. 
\end{example}

To address the inherent complexity of social challenges, we may also use FM-agents in the form of Large Language Model (LLM) to process and understand human instructions, queries, and feedback from stakeholders to alter the priorities within the resource allocation process \cite{sun2024adaplanner,wang2024ecg,xiao2024configurable,xu2024can}. For example, in the ARMMAN domain, a program manager may suggest prioritizing a specific segment of the underserved population such as those older in age, which an LLM could interpret and accordingly adjust the restless bandit resource allocation model by changing its reward function~\cite{behari2024decision,DBLP:journals/corr/abs-2408-12112}.
%The FM-agent should incorporate human instructions to iteratively improve upon the solution method design for the base-level problem. The FM-agent should tailor the solution method to the need of the local community being served. 

Besides the adaptation aspect, fostering efficient collaboration among multiple base-level agents plays an important role in AI4SI research. Recall that a base-level agent serves as an abstract representation of an individual or an entity in the real-world. In some problems, multiple base-level agents may collectively plan to counter an adversary such as wildlife poacher or terrorists \cite{shieh2012protect, gordon2024combining,xiong2024provably}. In other problems, multiple base-level agents may communicate to mitigate the impact of data errors, which frequently arise in real-world situations due to factors such as inconsistent data collection methods and deliberate noise introduced for differential privacy \cite{dubrow2022local, paulus2023reinforcing, dong2023symmetry, zhao2024bandit,dwork2014algorithmic, dong2022low}. 

However, previous works in AI4SI often require human experts in AI to manually craft ways of collaboration among base-level agents for specific AI4SI domains. An FM-agent can help with this process to accelerate AI for social impact work:

\begin{challenge}
The FM-agent, when designing solution methods for base-level systems, should design effective communication channels and strategies for base-level agents. Specifically, these channels should allow base-level agents to learn from each other's experience and to improve decision-making. 
\end{challenge}

Although the above vision on developing collaboration may appear to be straightforward for human-AI experts crafting solution methods, it is not easy for FM-agents to figure out due to complex relationships between base-level agents \cite{wang2020roma,wang2020rode,kang2022non}. The FM-agent may use the world knowledge of LLMs to understand which communications can be potentially useful and should be included in the design of solution methods \cite{chen2024internet}. Here a communication channel may be that an individual or entity talks to another via cellular network or other infrastructure in place. For example, the FM-agent may find that rangers, who work together to patrol protected areas such as a national park for illegal activities, may share their observations to improve the understanding of poaching hotspots and optimize the assignments for future patrols \cite{gordon2024combining}.

\subsection{Ethics and Fairness}
% In previous sections, we discussed visions that leverage Foundation models and LLMs. Advances in foundation models and LLMs have not been widely adapted in AI4SI, and little is known about their ethics and fairness aspect in high-stake social decision making \cite{anthis2024impossibility}
In high-stakes resource allocation scenarios like healthcare, authorities frequently prioritize certain groups based on sensitive attributes, aiming to address the needs of those most disadvantaged \cite{amon2020ending,vermagroup}. For example,  governments may mandate non-discrimination based on sensitive attributes, while non-profits may prioritize low-income groups. 
%A purely utilitarian approach might favor more privileged groups, potentially exacerbating socio-economic disparities. Furthermove, equal allocation is often not sufficient and equitable or balanced outcomes are preferred \cite{marsh1994equity,luss2012equitable}. Additionally, equal allocation of tasks amount social workers in non-profits is crucial to avoid excessive workload on social workers \cite{biswas2023fairness}. 
Given the importance of fairness in base-level system design and the solution method's tangible impact on people's lives, we propose the following idea:

\begin{challenge}
Ensure that the FM-agent recommends the design of a base-level system that does not discriminate against any subpopulation or result in unfavorable outcomes for under-privileged groups. 
\end{challenge}

When accelerating the design of base-level systems, the FM-agent should ensure fairness guarantees or fairness checks are in place. This can be done by explicitly incorporating fairness in designing base-level systems for social impact applications \cite{zehlike2017fa, vermagroup}. However, this added complexity is difficult for FM-agents to handle, due to the fact that AI may not easily understand demographic information available in text or abstract fairness concepts potentially based on demographics. Blindly applying fairness constraints or solely optimizing a fairness objective may greatly compromise the overall effectiveness of the solution method. Thus, fairness concerns necessitate innovative strategies to ensure that FM-agents design base-level systems that balance fairness and utility. 

\begin{example}
In ARMMAN a concrete example of fairness constraints is the enforcement of non-discrimination based on sensitive attributes and the prioritization of low-income and low-education groups to reduce socio-economic disparities. In the ARMMAN context, demographic information for enrolled beneficiaries are available. When designing a base-level system, the FM-agent could enforce fairness constraints such as that each beneficiary must receive a sufficient amount of resources within some time. Additionally, the FM-agent should prioritize underprivileged groups by explicitly optimizing a fairness objective (e.g. Max Nash Welfare or Maximin Reward) in the solution method.  
\end{example}

\section{Testing and Deployment}
\label{sec:phase3}
In this phase, we explain how to thoroughly evaluate and deploy AI models for social impact domains. We use FM-agents to improve model testing and facilitate real-world deployment. 

Deploying an AI model in real-world social impact domains without sufficient simulation studies may result in poor decision-making on crucial public resources. %Poor allocation of resources in social impact domains may result in a waste of scarce resources and even harm under-privileged groups. 
Thus, thoroughly testing and evaluating AI algorithms or trained models is an important aspect of accelerating AI for social impact. Based on above, we propose the following research idea:

\begin{challenge}
Employ FM-agents, based on LLMs, to simulate agents' behaviors. Here we wish to build a powerful simulator that serves as a good proxy of real-world deployment environment and can effectively evaluate the performance of trained models. 
\end{challenge}

LLM based simulators have recently received great interest, and there is demonstrated success in using LLMs to simulate human behaviors in fields including education, healthcare, and social sciences \cite{cheng2023compost,santurkar2023whose,argyle2023out,10.1145/3627673.3679071, markel2023gpteach}. To build such a LLM simulator / evaluator for AI4SI problems, we should represent observations and possible decisions in a way that the LLM can understand, and potentially use textual descriptions combined with structured data, to help the LLM simulator generate contextually appropriate (e.g. suitable for the domain) individual behaviors. 
%Textual descriptions on the domains, including meanings of the state space and the action space, would help the LLM to understand and generate contextually appropriate (e.g. suitable for the domain) agent behaviors or next states. 
Furthermore, textual descriptions on individual's characteristics, such as demographic information (age, gender, geographical location, etc) may help LLMs to better understand how individuals' condition would evolve over time. For a particular AI4SI problem, we may need to finetune LLMs on historical data collected to better simulate the individual's trajectories. %One may use one single LLM-based FM-agent to simulate the behaviors of multiple base-level agents, which would allow generalization across agents and is suitable when we have data for a large number of base-level agents but limited data for each base-level agent. When there is vast real-world data for each base-level agent, one may consider building a personalized simulator for each base-level agent. 

\begin{example}
In the ARMMAN application, we need to thoroughly test algorithms before real-world deployment. An FM-agent can employ LLMs to perform agent-based simulations to evaluate learning algorithms \cite{park2023generative, guo2024large}, or use cognitive models to augment ML based evaluation approaches \cite{gonzalez2003instance, seow2024improving}. 
\end{example}

Having discussed evaluation before real-world deployment, we now move on to challenges in the deployment. During the deployment of AI models, there can be shifts in the user base or shifts in people's behaviors \cite{elmachtoub2023balanced}. Different from adaptation ability taken into account during model development and training, distribution shifts in testing can be unexpected and the need to handle these shifts can be urgent. This brings the next research idea: 

\begin{challenge}
Have an FM-agent that could (i) involve human-in-the-loop and implement real-time monitoring to track model performance and detect shifts in user behavior or data distribution; (ii) use feedback from either human or AI to adjust the model (e.g. enhance model fairness when there are unexpected distribution shifts).  
\end{challenge}

Specifically, we may use a feedback loop to gather data on model predictions and user interactions, allowing for prompt detection of shifts in behavior \cite{behari2024decision,xiong2024dopldirectonlinepreference}. Once substantial shifts in user behaviors are detected, we could then involve human experts, potentially from partnering non-profit organizations, to review and provide feedback on model predictions and decisions. This feedback can then be used to guide model adjustments and improve its response to distribution changes. We may retrain or finetune the model using newly collected data, potentially weighting recent data more to better align with current trends and user behavior \cite{bommasani2021opportunities,choromanski2023efficient, peng2024tamba,kong2024aligning,zhaoself}.

\section*{Acknowledgments}
This work is supported by the Harvard Data Science initiative. 

%% The file named.bst is a bibliography style file for BibTeX 0.99c
\bibliographystyle{named}
\bibliography{ijcai24}

\end{document}